# Machine Learning Applications in Studying Mental Health Among Immigrants and Racial and Ethnic Minorities: A Systematic Review

Khushbu Khatri Park[1], Abdulaziz Ahmed, PhD[1*], Mohammed Ali Al-Garadi, PhD[2]

**[1]Department of Health Services Administration, School of Health Professions, University of Alabama at Birmingham, 1716 9th Ave S, Birmingham, AL 35233, USA**
**[2]Department of Biomedical Informatics, School of Medicine, Vanderbilt University, 1161 21st Ave S # D3300, Nashville, TN 37232, USA**

[*]Correspondence to: Dr. Abdulaziz Ahmed, Department of Health Services Administration, School of Health Professions, University of Alabama at Birmingham, Birmingham, 35233, USA, aahmed2@uab.edu

**Abstract**

Background

The use of machine learning (ML) in mental health (MH) research is increasing, especially as new, more complex data types become available to analyze. By systematically examining the published literature, this review aims to uncover potential gaps in the current use of ML to study MH in vulnerable populations of immigrants, refugees, migrants, and racial and ethnic minorities.

Methods

In this systematic review, we queried Google Scholar for ML-related terms, MH-related terms, and a population of a focus search term strung together with Boolean operators. Backward reference searching was also conducted. Included peer-reviewed studies reported using a method or application of ML in an MH context and focused on the populations of interest. We did not have date cutoffs. Publications were excluded if they were narrative or did not exclusively focus on a minority population from the respective country. Data including study context, the focus of mental healthcare, sample, data type, type of ML algorithm used, and algorithm performance was extracted from each.

Results

Our search strategies resulted in 67,410 listed articles from Google Scholar. Ultimately, 12 were included. All the articles were published within the last 6 years, and half of them studied populations within the US. Most reviewed studies used supervised learning to explain or predict MH outcomes. Some publications used up to 16 models to determine the best predictive power. Almost half of the included publications did not discuss their cross-validation method.

Conclusions

The included studies provide proof-of-concept for the potential use of ML algorithms to address MH concerns in these special populations, few as they may be. Our systematic review finds that the clinical application of these models for classifying and predicting MH disorders is still under development.

Funding: This research did not receive any specific grant from funding agencies in the public, commercial, or not-for-profit sectors.

## 1. Introduction

Common Mental Disorders (CMDs), including major depressive disorder, mood disorder, anxiety disorder, and alcohol use disorder, affect approximately one in five people worldwide.[1,2] More specifically, the global prevalence of post-traumatic stress symptoms is 24.1%, anxiety is 26.9%, sleep problems is 27.6%, depression is 28.0%, stress is 36.5%, and psychological distress is 50.0%.[3]

Although racial and ethnic minorities have chronic exposures to systematic disadvantages and discrimination, paradoxically, black and Latinx populations have similar or better rates of mental health (MH) than non-Hispanic whites for most major MH disorders, including major depression.[4-6] Furthermore, they have a lower lifetime prevalence of mood, anxiety, and substance use disorders than non-Hispanic whites.[7,8] Black and Latinx individuals, however, are at higher risk of persistence and disability from mental illness.[8-10] Asian Americans have the best MH status compared to whites and other racial and ethnic minorities, but this is poorly studied.[11] Ultimately, CMDs may disproportionately affect ethnic and racial minorities overrepresented in homeless, incarcerated, and medically underserved populations.[4]

The intersection of mental illness stigma, interpersonal and structural discrimination, and low socioeconomic status perpetuates the disparities in MH outcomes.[2,12] Moreover, because minority populations and individuals with low socioeconomic status are more likely to have limited health literacy, have geographic inaccessibility to MH services, and lack medical insurance, MH disparities are further widened.[13] Research shows that racial and ethnic minorities have a lower quality of MH services, are less satisfied with professional MH services, and have higher dropout rates than whites.[7,14-17]

Interestingly, immigrants tend to report having better MH than their counterparts in both their countries of origin and host countries, at least initially. This phenomenon is known as the "immigration paradox." Immigrants arrive healthier, but after facing the challenges and stresses of assimilation, their initial advantage deteriorates, and the healthy immigration effect (HIE) disappears.[18,19] Over time, the rates of mental illness worsen to match or exceed those of the general population, especially in psychotic disorders.[18-20] A study from a Canadian national database shows that recent immigrants (landed within ten years) reported having better MH than native Canadians, but after the adjustment period, they became more likely to report worse MH levels than their Canadian-born counterparts.[18]

Refugees differ from immigrants in fleeing war, persecution, and/or political turbulence, whereas immigrants leave their native country to settle permanently in a foreign country. When broken down by immigrant admission categories, recently arrived refugees reported lower rates of self-reported MH than other immigrant groups and native-born Canadians.[18] Asylum seekers and refugees are at substantially higher risk of psychiatric disorders and have up to 10 times the rate of post-traumatic stress disorder[20] due to exposure to expulsion factors such as war, poverty, and persecution.

Migrants experience unique MH stressors, resulting in different needs than native populations. The process of pre-migration, migration, and post-migration settlement requires changes in personal and social relationships and adopting new socioeconomic and cultural environments.[19,21,22] These major transitions, with exposures to stress, helplessness, and poor resources, significantly impact mental well-being. A sense of powerlessness during the pre-migration and migration phases can initiate or worsen CMDs.[23] However, during the post-migration resettlement phase, migrants may initially feel relief and success in reaching their destination, facilitating the HIE. Over time, the strain of rebuilding social networks while facing social alienation and reestablishing economic stability in the face of structural barriers and inequalities has negative consequences on MH status.[19,22,24] Immigrants face racism, discriminatory and exclusionary policies, status loss, and sometimes violence, which further affect their mental well-being.[19,25,26]

With more than 258 million people living outside their country of birth and millions more racial and ethnic minorities living amongst their majority counterparts, there is a need to understand and strengthen the MH resiliency of these populations. Clinicians and researchers have increasingly collected "big data" to aid this mission. This includes structured and unstructured data from electronic health records (EHR), smartphones, wearables, social media, and other large, complex sources. However, more robust data analysis methods are necessary to gain insights from this data. Machine learning (ML), the intersection of computer science, artificial intelligence, bioinformatics, and biostatics, offers methods to handle such data types (e.g., images, text, and tabular data) which cannot be handled by traditional analytical methods, like t-tests.[27] Broadly, ML encompasses the design and development of algorithms and statistical models that learn from data to make predictions of new data.

The use of ML in the health sciences is increasing. Several review articles discuss the current ML applications in MH and the development of better ML models in MH research but do not specifically include discussions of race, ethnicity, or immigration status.[28,29] Maslej et al.[30] conducted a rapid review and utilized a Critical Race Theory perspective to examine how race and racialization are defined in the application of ML in MH. However, their study focused on only Major Depressive Disorder and did not include other CMDs.

This systematic review asks: what is the potential of ML to study MH in vulnerable populations of immigrants, refugees, migrants, and racial and ethnic minorities? Moreover, how feasible is it, and what are the limits of ML in how it is currently being used in this context? By systematically examining the published literature through this perspective, we aimed to uncover potential biases and gaps in the literature. This review paper will guide researchers to develop more robust ML methods to study MH among vulnerable populations. Furthermore, this review will highlight the gaps in the field of ML applications in MH for these populations.

## 2. Methods

We employed the Preferred Reporting Items for Systematic Reviews and Meta-Analyses (PRISMA) guidelines.[31] Two reviewers (KP and AA) independently conducted exploratory searches in Google Scholar. All queries had three components: an ML-related term ("machine learning," "artificial intelligence," "deep learning"), an MH-related term ("mental health," "mental illness," "attention deficit hyperactivity disorder" (ADHD), anxiety, bipolar, depression, "post-traumatic stress disorder" (PTSD), schizophrenia) and a population of a focus search term (immigrant*, migrant*, refugee, race*, ethnic*, minority*, Latino, African American). These terms were combined with the Boolean "AND" operator to create a final search string. Table 1 lists some of these combinations. These search terms were conducted on titles, keywords, and abstracts. Backward reference searching was also conducted, reviewing references from the articles that matched our search criteria for more articles that could fit our inclusion criteria.

**Table 1. A sample of search term syntax**

| | | | | |
|---|---|---|---|---|
| "Machine learning" | AND | "Mental health" | AND | Race* OR Ethnic* |
| "Machine learning" | AND | "Mental health" | AND | Minority* |
| "Machine learning" | AND | "Mental health" | AND | Immigrant* OR Migrant* OR Refugee* |
| "Artificial intelligence" OR "Deep learning" | AND | "Mental health" | AND | Immigrant |
| "Machine learning" | AND | Depression | AND | Migrant |
| "Machine learning" | AND | Depression | AND | Immigrant |
| "Machine learning" | AND | Depression | AND | Refugee |

Inclusion criteria included: (i) the article reported using a method or application of ML in an MH context (ii) the primary population studied was immigrants, refugees, migrants, and/or racial and ethnic minorities (iii) the article was published in a peer-reviewed publication (iv) the article was available in English. We did not limit articles to just those published in America or have any date cutoffs. Articles were excluded if they were narrative in nature (e.g., commenting on future applications of ML in MH or did not add original contributions to ML in MH) or if they did not exclusively focus on a minority population from the respective country (e.g., a study of ethnically Chinese migrants in China would be excluded). Conflicts over inclusion were thoroughly discussed, and a consensus was sought before the inclusion or exclusion of the publication in question.

Data were extracted from each article, including study context, the focus of mental healthcare, sample, data type, type of ML algorithm used, and algorithm performance. A narrative synthesis approach was applied.

## 3. Results

To summarize the results of this review, we present them in three sections. The first section includes the results of the PRISMA selection process. The second section details the characteristics of the selected studies, such as their area of focus, publication year and location, and data source. The third section highlights the machine learning models used in the studies for predicting and studying mental health outcomes.

### *3.1 Selected studies*

Seven different permutations of the search terms were used to search Google Scholar. Our search strategies resulted in 67,410 listed articles from Google Scholar. Figure 1 shows the flow of our search strategy and results. The first 40 titles from each search were included for inclusion in this review for a total of 280. Based on titles and abstracts, 62 were selected and further reviewed. Most of these records were excluded because they did not focus on the population of interest. Instead, they focused on majority populations and racially homogenous populations and/or did not include discussions about immigrant/migrant status. We also reviewed five abstracts from citation searching. Ultimately 12 publications were included in this systematic review.

*3.2 Publication Characteristics*

Table 2 presents some high-level characteristics of the reviewed publications. All but two of the analyzed articles were published in the last three years, with two earliest from 2017.[32,33] Half of the papers were from the US or incorporated populations based in the US, four were from Europe, and the rest were from Asia. Among the 12 articles, five focused on refugee populations,[33-37], three focused on Hispanic populations in the US,[32,38,39] two focused on black individuals,[40,41], and the last two articles focused on Korean immigrants in the US[42] and immigrant populations in Europe.[43] The areas of mental health focus included stress,[36] ADHD,[40,41] trauma,[33,34] depression,[37,39,41] PTSD,[35] psychological distress,[42] schizophrenia,[43] suicidal ideation,[38] and substance abuse.[32]

**Table 2.** Publication analysis

| **Characteristic** | **N** | **%** | **Reference** |
|---|---|---|---|
| **Year of publication** | | | |
| 2017 | 2 | 16.7% | 32,33 |
| 2020 | 2 | 16.7% | 42 43 |
| 2021 | 5 | 41.7% | 36,37,39-41 |
| 2022 | 3 | 25.0% | 34,35,38 |
| **Region** | | | |
| Asia | | | |
| Turkey | 1 | 8.3% | 35 |
| Jordan | 1 | 8.3% | 34 |
| Europe | | | |
| United Kingdom | 1 | 8.3% | 37 |
| Germany | 2 | 16.7% | 33,36 |
| Switzerland | 1 | 8.3% | 43 |
| US | 6 | 50.0% | 32,38-42 |
| **Population of focus** | | | |
| Refugees | 5 | 41.7% | 33-37 |
| Hispanics | 3 | 25.0% | 32,38,39 |
| African Americans | 2 | 16.7% | 40,41 |
| Korean immigrants | 1 | 8.3% | 42 |
| European immigrants | 1 | 8.3% | 43 |

Surveys,[33,35,37,42] drawings,[34] secondary data sets (including EHR data and national sample sets), [32,34,38,43] internet-based posts,[36,39] and genomic sequencing data[40,41] were analyzed in the included publications (see Table 3). Various populations were considered, and sample sizes varied widely due to the type of data collected and analyzed. For example, Augsburger and Elbert [33] enrolled 56 resettled refugees in a study to prospectively analyze their risk-taking.[33] Goldstein, Bailey [38] used a retrospective dataset with 22,968 unique Hispanic patients, and Acion et al.[32] included 99,013 Hispanic individuals in their secondary data analysis. Children were also included in the reviewed studies; one examined the depression and PTSD levels of 631 refugee children residing in Turkey.[35] Another study analyzed drawings from 2480 Syrian refugee children to find the predictors of exposure to violence and mental well-being.[34] Other sample sets analyzed 0.15 million unique tweets from Twitter[36] and 441,000 unique conversations from internet message boards and social media sites.[39] Genomic sequencing data was collected from 4,179 black individuals[41] and 524 black individuals.[40]

Most reviewed studies used supervised learning intending to explain or predict certain MH outcomes. For example, to classify substance use disorder treatment success in Hispanic patients, Acion et al. compared 16 different ML models to an ensemble method they called "Super Learning".[32] Similarly, Huber et al. compared various ML algorithms, including decision trees, support vector machines, naïve Bayes, logistic regression, and K-nearest neighbor, to determine the model with the best predictive power for classifying Schizophrenia spectrum disorders in migrants.[43] Two studies explored the impact of trauma exposure on MH using ML.[33,34] Two studies utilized social media data to understand MH at a population-health level through ML algorithms.[36,39] All study aims are found in Table 3.

*3.3 Machine Learning Model Performance and Characteristics*

Table 4 outlines a summary of ML characteristics and model performance. This review found that all 12 included publications fell into three categories: classification,[32,36,38,40-43] regression,[33-35,37], and unsupervised topic modeling.[39]

The publications used a range of ML models, from one[33-35,37,40-42] to 16.[32] In studies where multiple ML models were used, the aim was often to compare the models to determine the best predictive power. For example, Acion et al. compared 16 models and evaluated them using the area under the receiver operating characteristic curve (AUC) to classify substance use disorder treatment success in Hispanic patients.[32] Similarly, Huber et al. compared five different ML algorithms, including decision trees, support vector machines, naïve Bayes, logistic regression, and K-nearest neighbor, to determine the model with the best predictive power for classifying Schizophrenia spectrum disorders in migrants.[43] Two of the studies used linear regression.[35,37] All of the studies developed custom models to meet their study aims. The most common programs used in these studies were R,[32,33] SPSS,[35,42] and Python.[36,40,41]

Predictors that were included in the modeling were sociodemographic characteristics,[32,35,38,42,43], and some also included MH variables and experiences[32,33,35,38,42,43] collected from EHRs or surveys. One study first determined which of the included 653 input variables (including sociodemographic data, childhood/adolescence experiences, psychiatric history, past criminal history, social and sexual functioning, hospitalization details, prison data, and psychopathological symptoms) were the best predictor variables and trained a final ML algorithm using only those.[43]

Two studies did not report the best algorithm performance.[38,39] For the other studies, accuracy and AUC were commonly reported. For example, Acion et al. classified substance use disorder treatment success in Hispanic patients and found that the AUC of studied models ranged from 0.793 to 0.820, with the ensemble method achieving an AUC of 0.820, which was not significantly better than the traditional logistic regression model's AUC of 0.805.[32] Huber et al. identified a tree algorithm that differentiated native Europeans and non-European migrants with schizophrenia with an accuracy of 74.5% and a predictive power of AUC = 0.75.[43] In Liu et al. [40], the trained ML model had an accuracy of 78% in predicting ADHD in African American patients.[40] In a similar study to classify ADHD, depression, anxiety, autism, intellectual disabilities, speech/language disorder, developmental delays, and oppositional defiant disorder in African Americans, the model had an accuracy of 65% in distinguishing patients with at least one MH diagnosis from controls.[41] A second prediction model aimed at predicting the diagnosis of two or more MH disorders had a low accuracy level, with an exact match rate of 7.2-9.3%.[41] Khatua and Nejdl [36] analyzed tweets acquired from Twitter feeds from self-identified refugees and categorized them into themes of the immigrant struggle with an accuracy of 61.61% and 75.89%.

The included studies also used *p* values to assess their ML algorithms. Goldstein and Bailey utilized multivariable logistical regression to examine the relationship between experienced discrimination and suicidal ideation in Hispanic patients.[38] They found that 19.0% of Hispanic patients who experienced discrimination also experienced suicidal ideation, compared to 11.5% of patients that did not experience discrimination ($p$=0.001). Moreover, Hispanic patients had 1.72 greater odds of having suicidal thoughts if they experienced discrimination compared to those that did not ($p$=0.003). A study by Erol and Seçinti used regression analysis to study the relationship between PTSD and depression and various predictors in adolescent refugee minors.[35] They found that moderate and severe changes in family income level and stress in food access predicted depression scores and PTSD symptoms ($p < 0.01$). Drydakis [37] used random effects models to estimate the relationship between the number of mobile applications that facilitate immigrants' societal integration and immigrants' integration, health, and mental health.[37] The results showed a negative association between the number of standard m-Integration applications and adverse MH status ($p < 0.01$). Accuracy was also measured using importance and normalized importance,[42] Root-mean-square error (RMSE),[33] and Least Absolute Shrinkage and Selection Operator (LASSO) coefficients.[34]

*3.4 Cross Validation*

Five studies used internal cross-validation methods.[32-34,41,43] Only one study used an external data set to validate their ML algorithm.[40] That external validation of the algorithm reduced the accuracy of their algorithm from 78% to 70-75%.[40] Almost half of the included publications did not use or discuss their cross-validation method.[35,37-39,42]

**Figure 1.** PRISMA flow chart for study selection

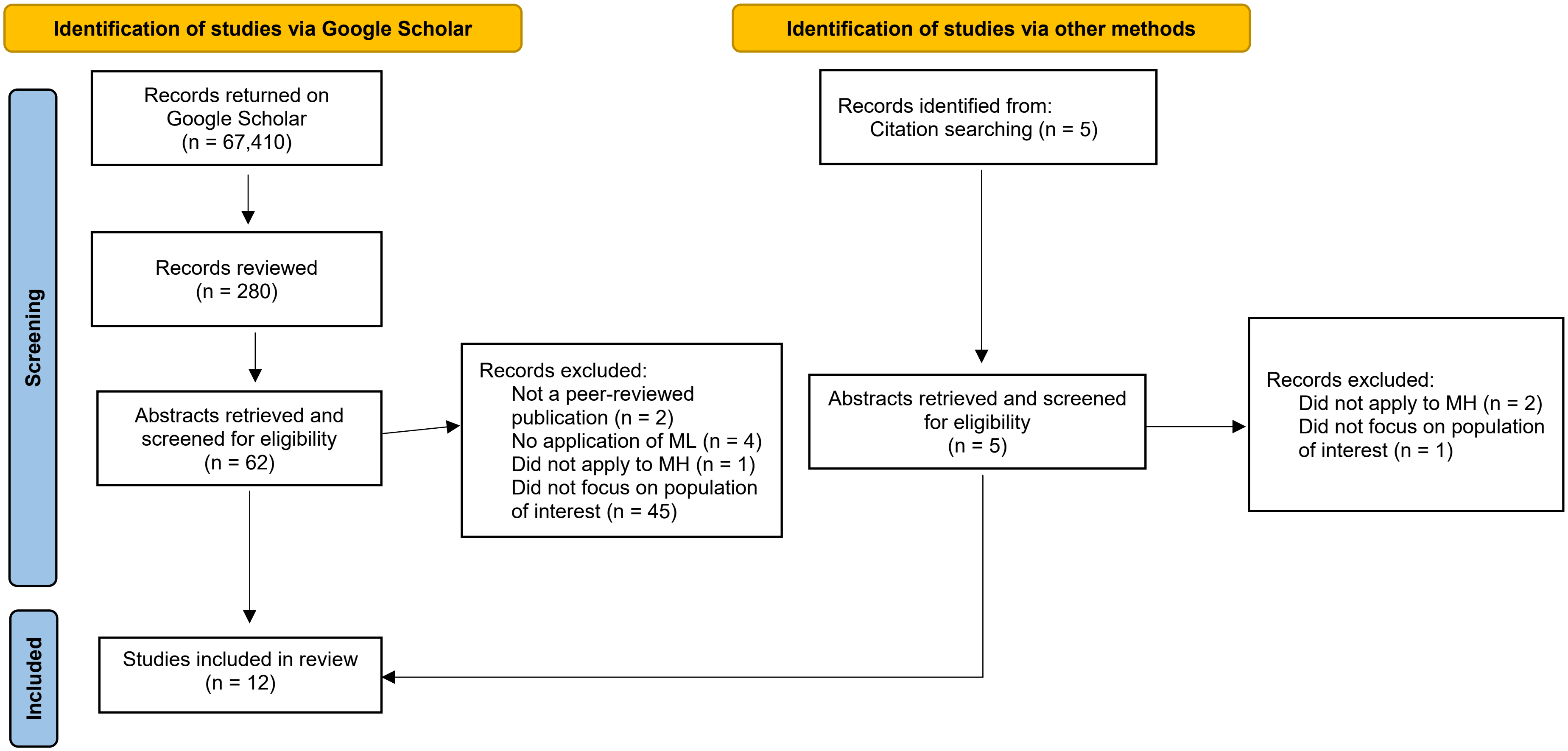

**Table 3.** Publication characteristics of included studies.

| First Author (year) | Study Aim | Area of MH focus | Sample size and characteristics | Data analyzed |
|---|---|---|---|---|
| Acion (2017) [32] | Predict substance abuse treatment success using 17 different machine learning models | Substance abuse | 99,013 Hispanic individuals | TEDS-D 2006-2011 |
| Augsburger (2017) [33] | Assessed risk-taking behavior in refugees after exposure to trauma using a gamified BART | Trauma | 56 Refugees resettled in Germany | Surveys and data on BART |
| Baird (2022) [34] | Used drawings by refugee children to estimate predictors of exposure to violence and mental wellbeing | Trauma | 2480 Syrian refugee children | USF 2016 dataset |
| Castilla-Puentes (2021) [39] | To understand how Hispanic populations converse about depression by conducting big data analysis of digital conversations through machine learning | Depression | 441,000 unique conversations about depression; 43,000 (9.8%) conversations were by Hispanics | Conversations from open sources like topical MH websites, message boards, social networks, and blogs |
| Choi (2020) [42] | Examined the predictive ability of discrimination-related variables, coping mechanisms, and sociodemographic factors on the psychological distress level of Korean immigrants in the U.S. during the pandemic | Psychological distress | 790 Korean immigrants, foreign and US-born | Surveys |
| Drydakis (2021) [37] | Understanding associations between the number of mobile applications in use aiming to facilitate immigrants' societal integration and increased level of integration, good overall health, and mental health | Depression | 287 immigrants in Greece | Surveys |
| Erol (2022) [35] | Examine the PTSD and depression levels of Syrian refugee children and adolescents, the difficulties they experienced in access to food and education, and the changes in their family income, and evaluate the effects of these factors on symptom severities of depression and PTSD | Depression & PTSD | 631 Refugee children living in Turkey | Surveys |
| Goldstein (2022) [38] | To examine the relationship between experiencing discrimination and suicidal ideation in Hispanic populations | Suicidal ideation | 22,968 Hispanics | Holmusk and MindLinc EHR datasets, 52,703 patient-year observations from 2010 to 2020 |
| Huber (2020) [43] | Differentiated native Europeans and migrants as to their risk of having schizophrenia | Schizophrenia | 370 patients with diagnosed schizophrenia spectrum disorder | Hospital data from 1982 to 2016 |
| Khatua (2021) [36] | Using social media data to identify the voices of migrants and refugees and analyze their MH concerns | General MH | 0.15 million tweets, 2% from self-identified refugees | 0.15 million tweets |
| Liu (2021) [41] | Used ML algorithms to distinguish ADHD, depression, anxiety, autism, intellectual disabilities, speech/language disorder, delays in | ADHD, depression, anxiety, autism, intellectual | 4179 black individuals | Genomic sequencing data |
| | development, and oppositional defiant disorder in blacks using the data from their genome. | disabilities, speech/language disorder, delays in development, oppositional defiant disorder | | |
| Liu (2021) [40] | Used ML algorithms to distinguish ADHD in blacks using the data from their genome. | ADHD | 524 black individuals | Genomic sequencing data |

Abbreviations: Balloon analogue risk task (BART), Treatment Episode Data Set-Discharge (TEDS-D)

**Table 4.** Machine Learning model characteristics from selected articles

| First Author (year) | Outcome Variable | Predictors (Input variables) | ML technique | Cross-validation method (internal, external) | Type | Program used | Best algorithm performance |
|---|---|---|---|---|---|---|---|
| Acion (2017) [32] | Substance abuse treatment success | 28; 10 patient characteristics, 3 treatment factors, referral type, problematic substance characteristics and mental health problem | LR, RLR, Lasso-LR, EN, RF, DNN, EL | Two-fold cross-validation (I) | Classification | R; H2O R interface and package rROC | AUC: 0.793 - 0.820<br>Best mode: EL |
| Augsburger (2017) [33] | Risk-taking behavior as measured using a balloon analog risk task (BART) | Exposure to different types of childhood maltreatment, experiences of war and torture, lifetime traumatic events and symptoms of depression | Stochastic GBM | Tenfold cross-validation with three repetitions (I) | Regression | R; *gbm* & *caret* | RMSE: 18.70, R^2: .20, |
| | | and PTSD, sociodemographic factors | | | | | |
| Baird (2022) [34] | Psychological trauma as measured on the GHQ-12 | 18 digitally coded features in self-portraits and free drawings | One model method used: LASSO-R | K-fold cross-validation (I) | Regression | Not reported | R-squared: 0.108 |
| Castilla-Puentes (2021) [39] | Tone, topics, and attitude of digital conversations | Digital conversations | NLP and texting mining | Not used | Unsupervised-Topic modeling | CulturIntel | Not reported |
| Choi (2020) [42] | Psychological distress is measured using the Kessler Psychological Distress Scale (K10) | Demographic characteristics, three types of discrimination characteristics, three types of coping mechanisms | ANN | Not used | Classification | SPSS | AUC: 0.806 |
| Drydakis (2021) [37] | Increased level of integration, overall health, and mental health | Number of mobile applications in use that facilitate immigrants' societal integration | Linear Regression | Not used | Regression | Not reported | $p < 0.005$ |
| Erol (2022) [35] | Symptom severity of depression and PTSD | Demographic data, PTSD and depression levels, access to food and education, and changes in family income | Linear regression | Not used | Regression | SPSS | R-squared = 0.123 |
| Goldstein (2022) [38] | Suicidal ideation in the past year | Experience of discrimination, demographics | Deep-learning NLP algorithms and LR | Not used | Classification | Not reported | Not reported |
| Huber (2020) [43] | Migrant status | 653 variables | LR, DTs, SVM, and naive Bayes | 5-fold cross-validation (I) | Classification | Not reported | DT Accuracy: 74.5%; AUC: 0.75 |
| Khatua (2021) [36] | Tweets that fall into 3 themes: generic views, initial struggles, and subsequent settlement | Tweets | Bi-LSTM, CNN, BERT | Training and testing | Classification | Python | F1-Score: 61.61 - 75.89% |
| Liu (2021) [41] | MH diagnosis from EHR | Copy number variation | Multi-layer perceptron | Two-fold random shuffle test validation (I) | Classification | Python; Scikit-learn package | Accuracy: 65.7% |
| Liu (2021) [40] | ADHD diagnosis | Copy number variation | Multi-layer perceptron | Two-fold random shuffle test validation (E) | Classification | Python; Scikit-learn package | Accuracy: 75.4% |

Abbreviations: Logistic regression (LR), Ridge logistic regression (RLR), Least Absolute Shrinkage and Selection Operator, (Lasso-LR), random forests (RF), deep learning neural networks (DNNs), Ensemble learning (EL), Lasso-Regression (Lasso-R), gradient boosting machines (GBM), Natural language processing (NLP), Artificial Neural Network (ANN), decision trees (DTs), support vector machines (SVM), Bidirectional Long Short-Term Memory (Bi-LSTM) and Convolutional neural network (CNN), Bidirectional Encoder Representations from Transformers (BERT), area under the receiver operating characteristic Curve (AUC), Root-mean-square error (RMSE)

## 4. Discussion

In recent years, there has been significant interest in the potential of ML to transform the field of MH research.[29] Studies examining ML models have generally concluded that they outperform traditional statistical models, creating exciting possibilities for applying ML to studying MH in various populations. Recent advances in computational power and software availability have enabled researchers to reach new audiences and demonstrate the clinical value of ML. In particular, some studies have aimed to inform clinicians about the methods and applications of ML in the context of psychotherapy.[44] However, while many of the reviewed papers provide proof-of-concept for the potential use of ML algorithms to address MH concerns, our systematic review finds that the clinical application of these models for classifying and predicting MH disorders is still under development.

Despite ML's great interest and potential to transform MH research, few researchers have focused on specific and marginalized populations. In reviewing hundreds of articles on MH and ML, we found only a handful of studies specifically targeting immigrants, migrants, refugees, and/or racial and ethnic minorities. Many researchers simply included race as a variable in their models rather than designing ML algorithms to analyze these specific groups of individuals. [45,46] Moreover, as noted by Maslej et al., [30] most studies that considered African American and white samples used self-reported race or ethnicity or did not describe how this information was collected and thus were excluded from our analysis.

The current lack of ML models tailored to specific populations presents opportunities and challenges. On the one hand, it can help prevent the perpetuation of health disparities that arise when models built on majority populations are used to misclassify minorities.[47] Performance differences in ML exist for different populations, especially with genomic data. For instance, one study externally validated their algorithm[40] on white Americans rather than African Americans and found that their algorithm's accuracy decreased. On the other hand, this lack of tailored models highlights the opportunity for researchers and clinicians to bridge the gap between what is known about majority populations and what is yet to be uncovered in other populations. Training ML models on other groups could expedite this process without being too resource intensive.

One of the most common challenges in utilizing ML techniques to build classifiers for MH is the use of small sample sizes, which may limit the representation of the entire population and impact the generalizability of the classifier's accuracy estimate. This can be a practical limitation due to resource constraints in real-world clinical or diagnostic settings. However, researchers need to understand that using ML alone cannot address this issue.[48] Most ML methods rely on supervised learning models, which are successful due to the abundance of training data. However, this training data requires human annotation, which can be time-consuming and costly. In the case of MH, there are insufficient publicly annotated datasets, making the quality of the data a significant concern for developing reliable models.[49] Another challenge of using ML for behavioral diagnosis is validating the classification algorithms against questionnaires or clinical diagnoses, which are known to have self-report biases and are not completely accurate. This highlights the lack of established best standards in the diagnosis process for mental disorders and other psychiatric conditions.[50]

Future directions include the development of more robust and generalizable algorithms that can improve prediction capabilities. ML can be leveraged to understand the prevalence of MH conditions at a population level by using open-source and freely available data, which can be more accurate and less labor-intensive than traditional surveys. Furthermore, ML can enable the study of MH in children and adolescents in innovative ways. [34,41] The application of these models can be expanded to different sources and sample sizes, potentially leading to a rapid increase in their use in clinical settings.

There is also potential for future application of ML and natural language processing (NLP) approaches to infer psychological well-being and detect CMDs in marginalized individuals based on their social media posts on platforms like Facebook and Twitter. Researchers must implement diagnostic criteria and tools that are precise and suitable for various online populations. Collecting personal information, such as sociodemographic characteristics and behavioral aspects, must be done in accordance with ethical considerations. These inferences can create online platforms that provide health information, support, and tailored interventions. Currently, the computational techniques and evaluations employed for collecting, processing, and utilizing online written data remain scattered throughout academic literature. [51] Moreover, this potential is limited by factors such as class imbalance, noisy labels, and text samples that are either too long or too short, which can lead to performance and stability issues. The diversity of writing styles and semantic heterogeneity in different data sources can also cause a lack of robustness in model performance. Standardizing these measures can allow for the development of scalable approaches for automated monitoring of public psychological health in the future. [52]

This review had limitations, including the possibility of missing relevant studies due to specificity in search terms. Future studies should consider using broader search terms to address these limitations. Additionally, the ethical and social implications of using ML in MH, including the potential for perpetuating existing biases and social determinants of health, should be carefully considered. Discussing ethical concerns is important when utilizing textual data related to MH, given the significance of privacy and security of personal information, particularly health data.

## 5. Conclusions

In conclusion, ML can potentially transform how we understand CMDs, particularly among vulnerable populations. Immigrants and refugees face unique challenges related to migration and resettlement that can negatively impact their MH status, including poverty, discrimination, and exposure to trauma. Meanwhile, African Americans and Hispanics in the US are overrepresented in marginalized populations and have higher persistence and disability from mental illness. This review has found that, to date, few studies have used ML to predict and classify MH in these populations, despite the wide gap in health disparities that persist in accessing quality MH services and outcomes. The use of big data and ML algorithms in the health sciences is increasing and holds promise, but more study of ML applications in MH is warranted.

## 6. Summary Table

What is already known on the topic:

- Mental health disorders, like major depressive disorder, mood disorder, anxiety disorder and alcohol use disorder, burden up to 20% of the world's population.
- Some groups of people, like immigrants and refugees and racial/ethnic minorities suffer from MH health disparities.
- Machine learning models are increasingly available to study rich and complex sets of data.

What this study adds:

- This study found 12 examples of how ML is being utilized to understand MH in immigrants and refugees and racial/ethnic minorities.
- This study highlights the need for more specific research in this population to continue to understand ML bias.

**Contributors**

All authors edited and reviewed the final manuscript. All authors have read and agreed to the final version of the manuscript and to the decision to submit it. All authors had access to all the data. KP and AA have verified the data.

**Declaration of interests**

We declare no competing interests.

**Data Sharing**

Template data collection forms and the data extracted from included studies can all be made available upon request to Dr. Abdulaziz Ahmed.